\newif\ifJOURNAL
\JOURNALfalse
\newif\ifCONF
\CONFfalse
\newif\ifarXiv
\arXivfalse
\newif\ifWP
\WPfalse
\newif\ifFULL
\FULLfalse

\arXivtrue

\newif\ifnotCONF	
\notCONFtrue
\ifCONF\notCONFfalse\fi

\newif\ifnotarXiv	
\notarXivtrue
\ifarXiv\notarXivfalse\fi

\newif\ifTR		
\TRfalse
\ifarXiv\TRtrue\fi
\ifWP\TRtrue\fi
\newif\ifnotTR
\notTRtrue
\ifarXiv\notTRfalse\fi
\ifWP\notTRfalse\fi

\newlength{\picturewidth}
\ifarXiv
  \documentclass{article}
  \usepackage{amsmath,amsthm,amsfonts,amssymb,latexsym,graphicx}
  \newcommand{\Extra}[1]{}
\fi

\ifWP
  \documentclass{article}
  \usepackage{amsmath,amsthm,amsfonts,amssymb,latexsym,epsfig,graphicx}
  \input{OT2enc.def}
  
  \usepackage{CJK}
  \input{/Doc/Computing/Latex/kp.txt}
  \newcommand{\Extra}[1]{}
\fi

\ifFULL
  \documentclass{article}
  \usepackage{amsmath,amsthm,amsfonts,amssymb,latexsym,color,graphicx}
  \newcommand{\Extra}[1]{\red{#1}}
  \newcommand{\red}[1]{\textcolor{red}{#1}}
  
  \newcommand{\bluebegin}{\begingroup\color{blue}}
  \newcommand{\blueend}{\endgroup}

\fi

\allowdisplaybreaks[4]

\DeclareMathOperator{\Prob}{\mathbb{P}}

\ifnotCONF
  
  \newtheorem{proposition}{Proposition}

\fi

\title{Cross-conformal predictors}

\ifarXiv
\author{Vladimir Vovk\\
\texttt{v.vovk{\rm@}rhul.ac.uk}\\
\texttt{http://vovk.net}}
\fi

\ifWP
\author{Vladimir Vovk}

\fi

\ifFULL
\author{Vladimir Vovk\\
\texttt{vovk{\rm@}cs.rhul.ac.uk}\\
\texttt{http://vovk.net}}
\fi

\begin{document}
\maketitle

\begin{abstract}
  This note introduces the method of cross-conformal prediction,
  which is a hybrid of the methods of inductive conformal prediction and cross-validation,
  and studies its validity and predictive efficiency empirically.
\end{abstract}

\section{Introduction}
\label{sec:introduction}

The method of conformal prediction produces set predictions
that are automatically valid in the sense that their unconditional coverage probability
is equal to or exceeds a preset confidence level (\cite{vovk/etal:2005book}, Chapter~2).
A more computationally efficient method of this kind is that of inductive conformal prediction
(\cite{papadopoulos/etal:2002ICMLA}, \cite{vovk/etal:2005book}, Section~4.1,
\cite{lei/etal:2012-full}).
However, inductive conformal predictors are typically less predictively efficient,
in the sense of producing larger prediction sets as compared with conformal predictors.
Motivated by the method of cross-validation \cite{mosteller/tukey:1968,stone:1974},
this note explores a hybrid method, which we call cross-conformal prediction.

We are mainly interested in the problems of classification and regression,
in which we are given a training set consisting of examples,
each example consisting of an object and a label,
and asked to predict the label of a new test object;
in the problem of classification labels are elements of a given finite set,
and in the problem of regression labels are real numbers.
If we are asked to predict labels for more than one test objects,
the same prediction procedure can be applied to each test object separately.
In this introductory section and in our empirical studies
we consider the problem of binary classification,
in which labels can take only two values, which we will encode as 0 and 1.
We always assume that the examples (both the training examples
and the test examples, consisting of given objects and unknown labels)
are generated independently from the same probability measure;
this assumption will be called the \emph{assumption of randomness}.

The idea of conformal prediction is to try the two different labels, 0 and 1,
for the test object,
and for either postulated label to test the assumption of randomness
by checking how well the test example conforms to the training set;
the output of the procedure is the corresponding p-values $p^0$ and $p^1$.
Two standard ways to package the pair $(p_0,p_1)$ are:
\begin{itemize}
\item
  Report the \emph{confidence} $1-\min(p^0,p^1)$ and \emph{credibility} $\max(p^0,p^1)$.
\item
  For a given significance level $\epsilon\in(0,1)$
  output the corresponding prediction set $\{y\mid p^y>\epsilon\}$.
\end{itemize}

In inductive conformal prediction the training set is split into two parts,
the proper training set and the calibration set.
The two p-values $p^0$ and $p^1$ are computed by checking how well the test example
conforms to the calibration set.
The way of checking conformity is based on a prediction rule found from the proper training set
and produces, for each example in the calibration set and for the test example,
the corresponding ``conformity score''.
The conformity score of the test example is then calibrated to the conformity scores
of the calibration set to obtain the p-value.
For details, see Section~\ref{sec:ICP}.

Inductive conformal predictors are usually much more computationally efficient
than the corresponding conformal predictors.
However, they are less predictively efficient:
they use only the proper training set when developing the prediction rule
and only the calibration set when calibrating the conformity score of the test example,
whereas conformal predictors use the full training set for both purposes.

Cross-conformal prediction modifies inductive conformal prediction
in order to use the full training set for calibration
and significant parts of the training set (such as $80\%$ or $90\%$)
for developing prediction rules.
The training set is split into $K$ folds of equal (or almost equal) size.
For each $k=1,\ldots,K$ we construct a separate inductive conformal predictor
using the $k$th fold as the calibration set
and the rest of the training set as the proper training set.
Let $(p^0_k,p^1_k)$ be the corresponding p-values.
Next the two sets of p-values, $p^0_k$ and $p^1_k$,
are merged into combined p-values $p^0$ and $p^1$,
which are the result of the procedure.

In Appendix~\ref{sec:CCP1} we consider the most standard way of combining p-values, Fisher's method.
However, the method produces badly miscalibrated results
as it assumes the independence of the p-values being combined,
whereas in fact these p-values are heavily dependent.
In the main part of the note, namely in Section~\ref{sec:CCP2},
we, essentially, combine p-values by averaging them.
This leads to much better calibration;
since we have no theoretical results about the validity of cross-conformal prediction in this note,
we rely on empirical studies involving the standard \texttt{Spambase} data set.
Finally, we use the same data set to demonstrate the efficiency of cross-conformal predictors
as compared with inductive conformal predictors.
Section~\ref{sec:conclusion} states an open problem.

\section{Inductive conformal predictors}
\label{sec:ICP}

We fix two measurable spaces:
$\mathbf{X}$, called the \emph{object space}, and $\mathbf{Y}$, called the \emph{label space}.
The Cartesian product $\mathbf{Z}:=\mathbf{X}\times\mathbf{Y}$ is the \emph{example space}.
A \emph{training set} is a sequence $(z_1,\ldots,z_l)\in\mathbf{Z}^l$ of \emph{examples} $z_i=(x_i,y_i)$,
where $x_i\in\mathbf{X}$ are the \emph{objects} and $y_i\in\mathbf{Y}$ are the \emph{labels}.
For $S\subseteq\{1,\ldots,l\}$, we let $z_S$ stand for the sequence $(z_{s_1},\ldots,z_{s_n})$,
where $s_1,\ldots,s_n$ is the sequence of all elements of $S$ listed in the increasing order
(so that $n:=\left|S\right|$).

In the method of inductive conformal prediction,
we split the training set into two non-empty parts,
the \emph{proper training set} $z_T$ and the \emph{calibration set} $z_C$,
where $(T,C)$ is a partition of $\{1,\ldots,l\}$.
An \emph{inductive conformity measure} is a measurable function $A:\mathbf{Z}^*\times\mathbf{Z}\to\mathbb{R}$
(we are interested in the case where $A(\zeta,z)$
does not depend on the order of the elements of $\zeta\in\mathbf{Z}^*$).
The idea behind the \emph{conformity score} $A(z_T,z)$
is that it should measure how well the example $z$ conforms to the proper training set $z_T$.
A standard choice is
\begin{equation}\label{eq:score}
  A(z_T,(x,y))
  :=
  \Delta(y,f(x)),
\end{equation}
where $f:\mathbf{X}\to\mathbf{Y}'$ is a prediction rule found from $z_T$
as the training set
and $\Delta:\mathbf{Y}\times\mathbf{Y}'\to\mathbb{R}$ is a measure of similarity between a label and a prediction.
Allowing $\mathbf{Y}'$ to be different from $\mathbf{Y}$
(usually $\mathbf{Y}'\supset\mathbf{Y}$)
may be useful when the underlying prediction method gives additional information
to the predicted label;
e.g., the MART procedure used in Section~\ref{sec:CCP2} and Appendix~\ref{sec:CCP1}
gives the logit of the predicted probability that the label is $1$.

The \emph{inductive conformal predictor} (ICP) corresponding to $A$
is defined as the set predictor
\begin{equation}\label{eq:ICP}
  \Gamma^{\epsilon}(z_1,\ldots,z_l,x)
  :=
  \{y \mid p^y>\epsilon\},
\end{equation}
where $\epsilon\in(0,1)$ is the chosen \emph{significance level}
($1-\epsilon$ is known as the \emph{confidence level}),
the \emph{p-values} $p^y$, $y\in\mathbf{Y}$, are defined by
\begin{equation*} 
  p^y
  :=
  \frac
  {
    \left|\left\{
      i\in C \mid \alpha_i\le\alpha^y
    \right\}\right|
    +
    1
  }
  {\left|C\right|+1},
\end{equation*}
and
\begin{equation}\label{eq:alphas-ICP}
  \alpha_i
  :=
  A(z_T,z_i),
  \quad
  i\in C,
  \qquad
  \alpha^y
  :=
  A(z_T,(x,y))
\end{equation}
are the conformity scores.
Given the training set and a test object $x$ the ICP predicts its label $y$;
it \emph{makes an error} if $y\notin\Gamma^{\epsilon}(z_1,\ldots,z_l,x)$.

The random variables whose realizations are $x_i$, $y_i$, $z_i$, $x$, $y$, $z$
will be denoted by the corresponding upper case letters
($X_i$, $Y_i$, $Z_i$, $X$, $Y$, $Z$, respectively).
The following proposition of validity is almost obvious.

\begin{proposition}[\cite{vovk/etal:2005book}, Proposition~4.1]
  \label{prop:validity-ICP}
  If random examples $Z_1,\ldots,Z_l,$ $Z=(X,Y)$ are i.i.d.,
  the probability of error $Y\notin\Gamma^{\epsilon}(Z_1,\ldots,Z_l,X)$
  does not exceed $\epsilon$ for any $\epsilon$ and any inductive conformal predictor $\Gamma$.
\end{proposition}

The family of prediction sets $\Gamma^{\epsilon}(z_1,\ldots,z_l,x)$, $\epsilon\in(0,1)$,
is just one possible way of packaging the p-values $p^y$.
Another way, already discussed in Section~\ref{sec:introduction} in the context of binary classification,
is as the \emph{confidence} $1-p$, where $p$ is the second largest p-value among $p^y$,
and the \emph{credibility} $\max_y p^y$.
In the case of binary classification
confidence and credibility carry the same information
as the full set $\{p^y\mid y\in\mathbf{Y}\}$ of p-values,
but this is not true in general.

In our experiments reported in the next section we split the training set
into the proper training set and the calibration set in proportion $2:1$.
This is the most standard proportion
(cf.\ \cite{hastie/etal:2009}, p.~222, where the validation set
plays a similar role to our calibration set),
but the ideal proportion depends on the learning curve for the given problem of prediction
(cf.\ \cite{hastie/etal:2009}, Figure~7.8).
Too small a calibration set leads to a high variance of confidence
(since calibrating conformity scores becomes unreliable)
and too small a proper training set leads to a downward bias in confidence
(conformity scores based on a small proper training set cannot produce confident predictions).
In the next section we will see that using cross-conformal predictors
improves both bias and variance
(cf.\ Table~{\ref{tab:statistics}).

\section{Cross-conformal predictors}
\label{sec:CCP2}

\emph{Cross-conformal predictors} (CCP) are defined as follows.
The training set is split into $K$ non-empty subsets (\emph{folds})
$z_{S_k}$, $k=1,\ldots,K$,
where $K\in\{2,3,\ldots\}$ is a parameter of the algorithm
and $(S_1,\ldots,S_K)$ is a partition of $\{1,\ldots,l\}$.
For each $k\in\{1,\ldots,K\}$ and each potential label $y\in\mathbf{Y}$ of $x$
find the conformity scores of the examples in $z_{S_k}$ and of $(x,y)$ by
\begin{equation}\label{eq:alphas-CCP}
  \alpha_{i,k} := A(z_{S_{-k}},z_i),
  \quad
  i\in S_k,
  \qquad
  \alpha^y_k := A(z_{S_{-k}},(x,y)),
\end{equation}
where $S_{-k}:=\cup_{j\ne k}S_j$
and $A$ is a given inductive conformity measure.
The corresponding p-values are defined by
\begin{equation}\label{eq:p-CCP}
  p^y
  :=
  \frac
  {\sum_{k=1}^K\left|\left\{i\in S_k\mid\alpha_{i,k}\le\alpha^y_k\right\}\right|+1}
  {l+1}.
\end{equation}
Confidence and credibility are now defined as before;
the set predictor $\Gamma^{\epsilon}$ is also defined as before, by (\ref{eq:ICP}),
where $\epsilon>0$ is another parameter.

The definition of CCPs parallels that of ICPs,
except that we now use the whole training set for calibration.
The conformity scores (\ref{eq:alphas-CCP}) are computed as in (\ref{eq:alphas-ICP})
but using the union of all the folds except for the current one
as the proper training set.
Calibration (\ref{eq:p-CCP}) is done by combining the ranks of the test example $(x,y)$
with a postulated label in all the folds.

If we define the separate p-value
\begin{equation}\label{eq:p-CCP-separate}
  p^y_k
  :=
  \frac
  {\left|\left\{i\in S_k\mid\alpha_{i,k}\le\alpha^y_k\right\}\right|+1}
  {\left|S_k\right|+1}
\end{equation}
for each fold,
we can see that $p^y$ is essentially the average of $p^y_k$.
In particular, if each fold has the same size,
$\left|S_1\right|=\cdots=\left|S_K\right|$,
a simple calculation gives
\begin{equation}\label{eq:modified-mean}
  p^y
  =
  \bar p^y
  +
  \frac{K-1}{l+1}
  \left(
    \bar p^y - 1
  \right)
  \approx
  \bar p^y,
\end{equation}
where
$
  \bar p^y
  :=
  \frac1K
  \sum_{k=1}^K
  p^y_k
$
is the arithmetic mean of $p^y_k$
and the $\approx$ assumes $K\ll l$.

In this note we give calibration plots for 5-fold and 10-fold cross-conformal prediction.
We take $K\in\{5,10\}$ following the advice in \cite{hastie/etal:2009}
(who refer to Breiman and Spector \cite{breiman/spector:1992} and Kohavi \cite{kohavi:1995}).
In our experiments we use the popular \texttt{Spambase} data set.
The size of the data set is 4601, and there are two labels:
\texttt{spam}, encoded as 1, and \texttt{email}, encoded as 0.

We consider the conformity measure (\ref{eq:score}) where $f$ is output by MART
(\cite{hastie/etal:2009}, Chapter~10) and
\begin{equation}\label{eq:Delta}
  \Delta(y,f(x))
  :=
  \begin{cases}
    f(x) & \text{if $y=1$}\\
    -f(x) & \text{if $y=0$}.
  \end{cases}
\end{equation}
MART's output $f(x)$ models the log-odds of \texttt{spam} vs \texttt{email},
\begin{equation*} 
  f(x)
  =
  \log\frac
  {P(1\mid x)}
  {P(0\mid x)},
\end{equation*}
which makes the interpretation of (\ref{eq:Delta}) as conformity score very natural.
(MART is known \cite{hastie/etal:2009} to give good results on the \texttt{Spambase} dataset.)

\ifnotCONF
  The R programs used in the experiments described in this section and the appendix
  have been uploaded to \texttt{arXiv}.
  The programs use the \texttt{gbm} package
  with virtually all parameters set to the default values
  (given in the description provided in response to \verb'help("gbm")').
  The only parameter that has been modified is \verb"n.trees", the number of trees,
  which should be as large as possible
  and whose default value was clearly insufficient.
\fi

Figure~\ref{fig:calibration_CCP2} gives the calibration plots for the CCP and for 8 random splits of the data set
into a training set of size 3600 and a test set of size 1001
and of the training set into 5 or 10 folds.
There is a further source of randomness as the MART procedure is itself randomized.
The functions plotted in Figure~\ref{fig:calibration_CCP2} map each significance level $\epsilon$
to the percentage of erroneous predictions made by the set predictor $\Gamma^{\epsilon}$ on the test set.
Visually, the plots are well-calibrated (close to the bisector of the first quadrant).

\begin{figure}[tb]
\begin{center}
  \includegraphics[width=0.45\textwidth]{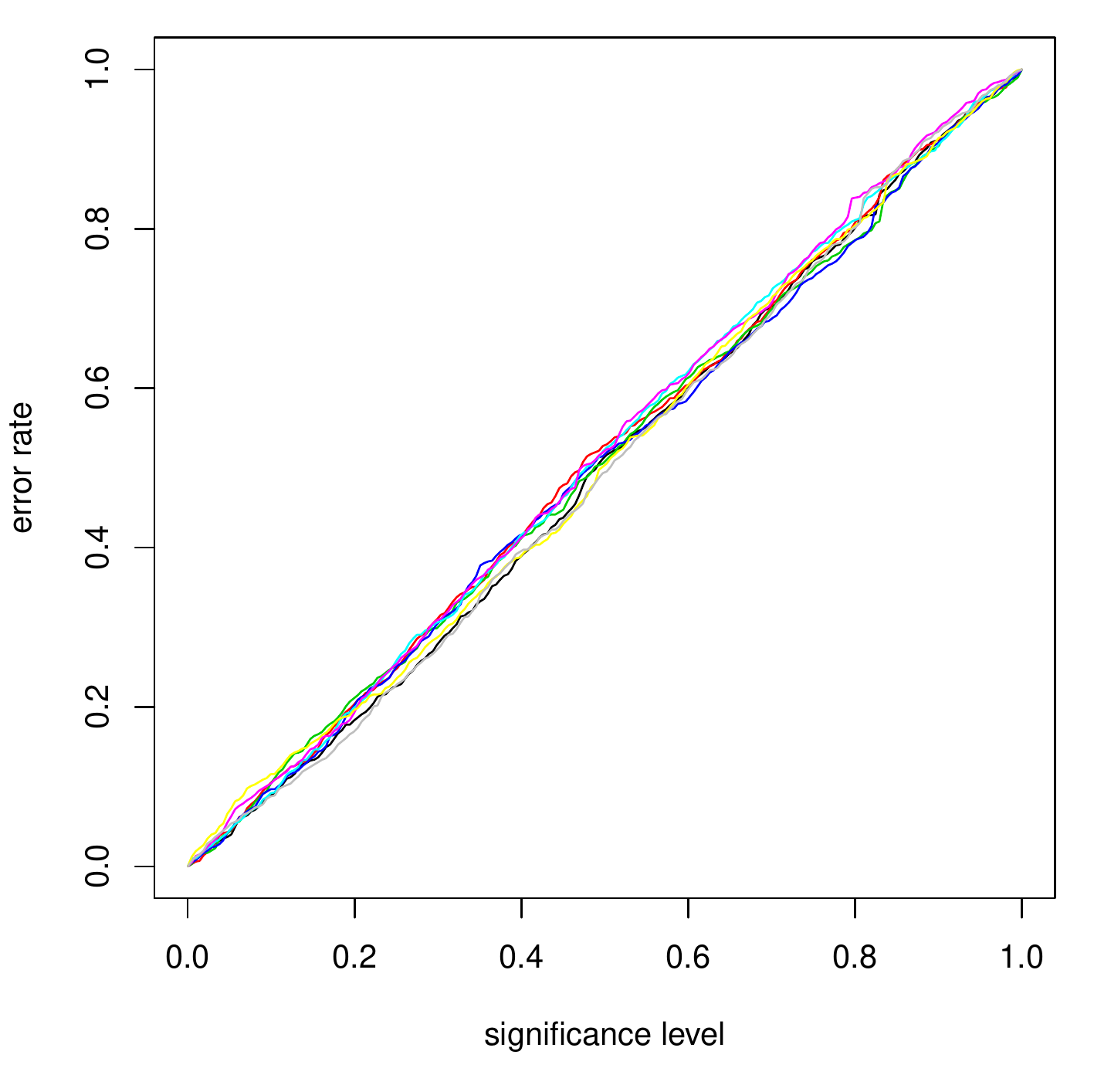}
  \includegraphics[width=0.45\textwidth]{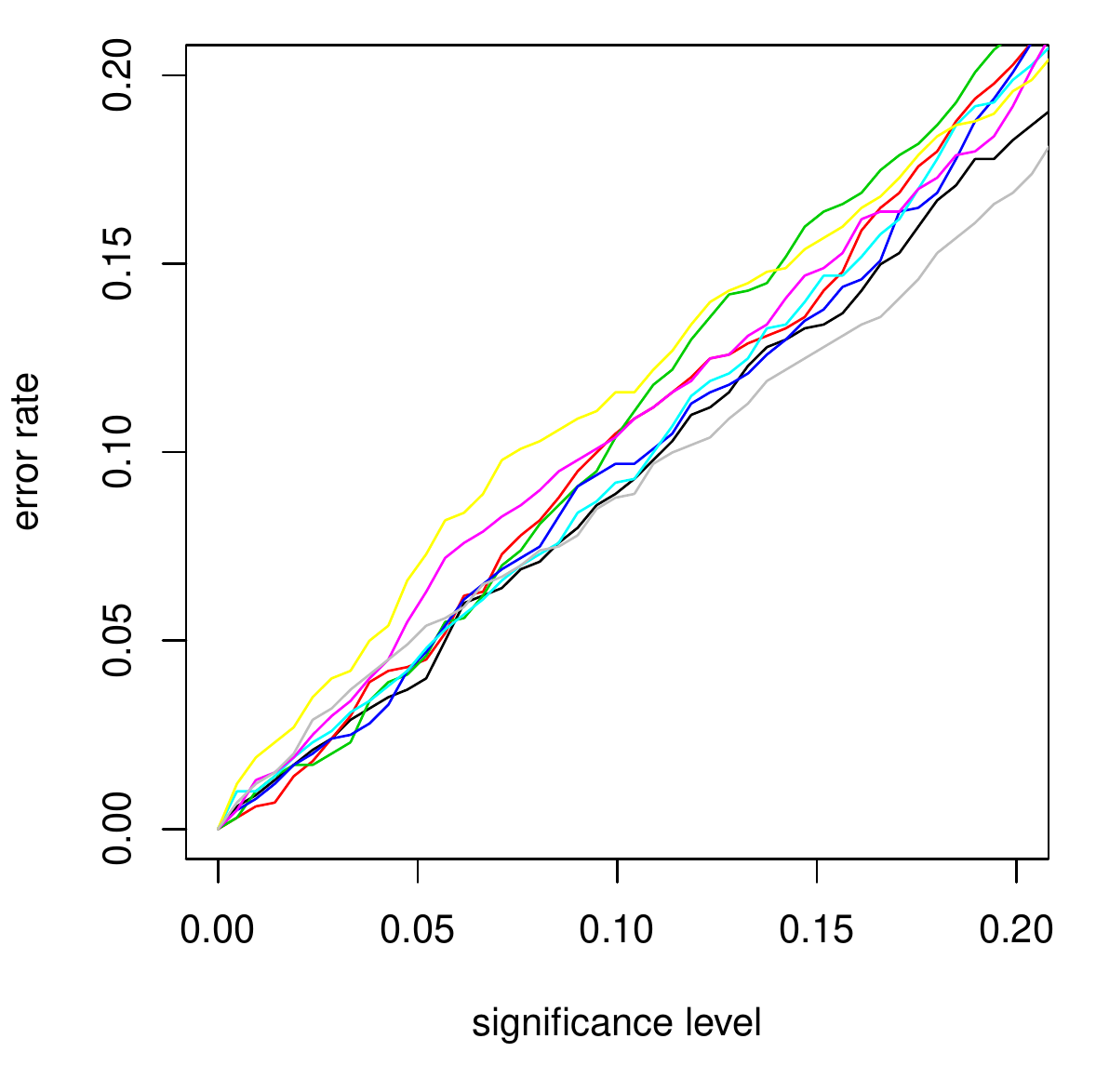}\\[5mm]
  \includegraphics[width=0.45\textwidth]{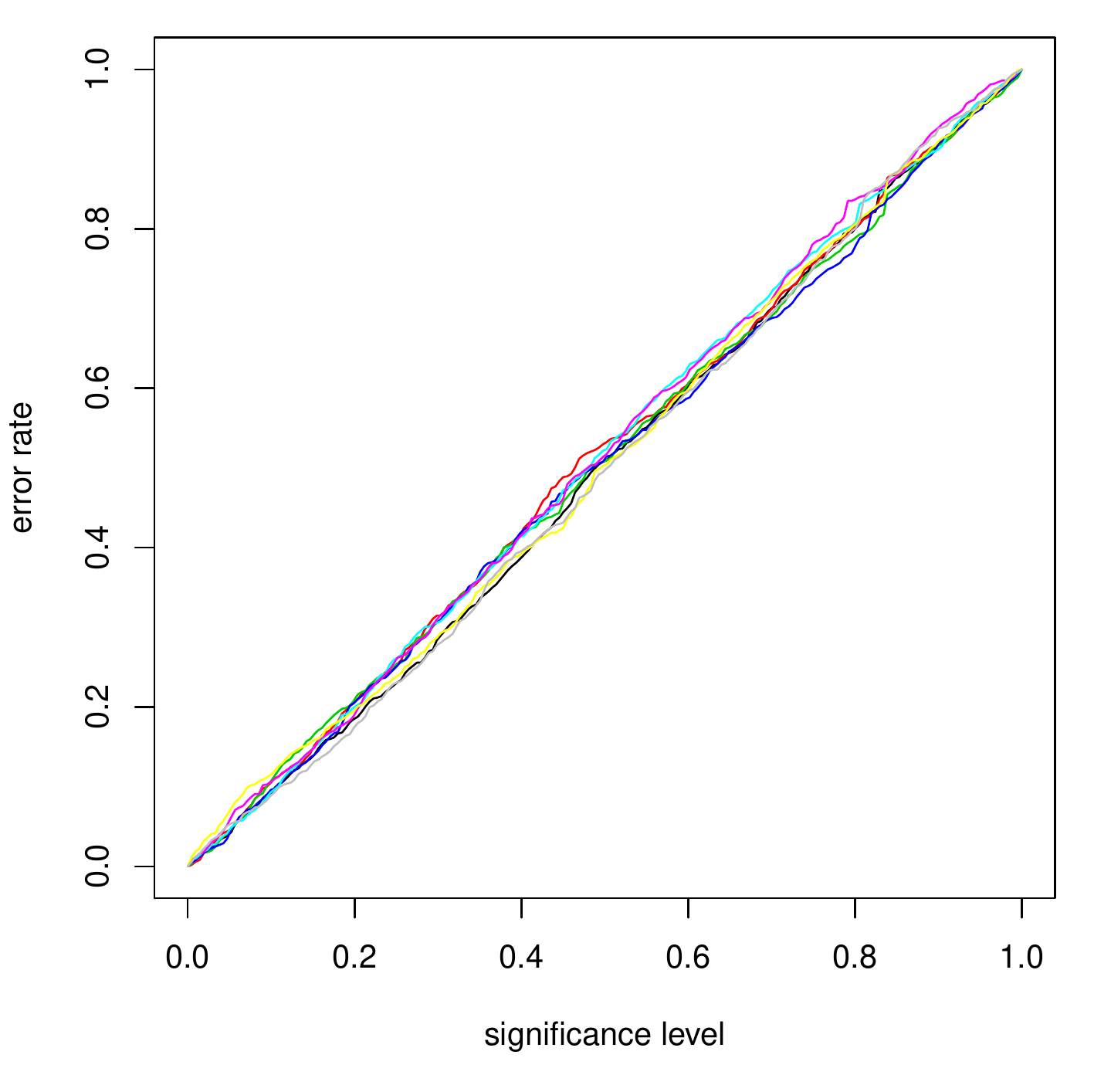}
  \includegraphics[width=0.45\textwidth]{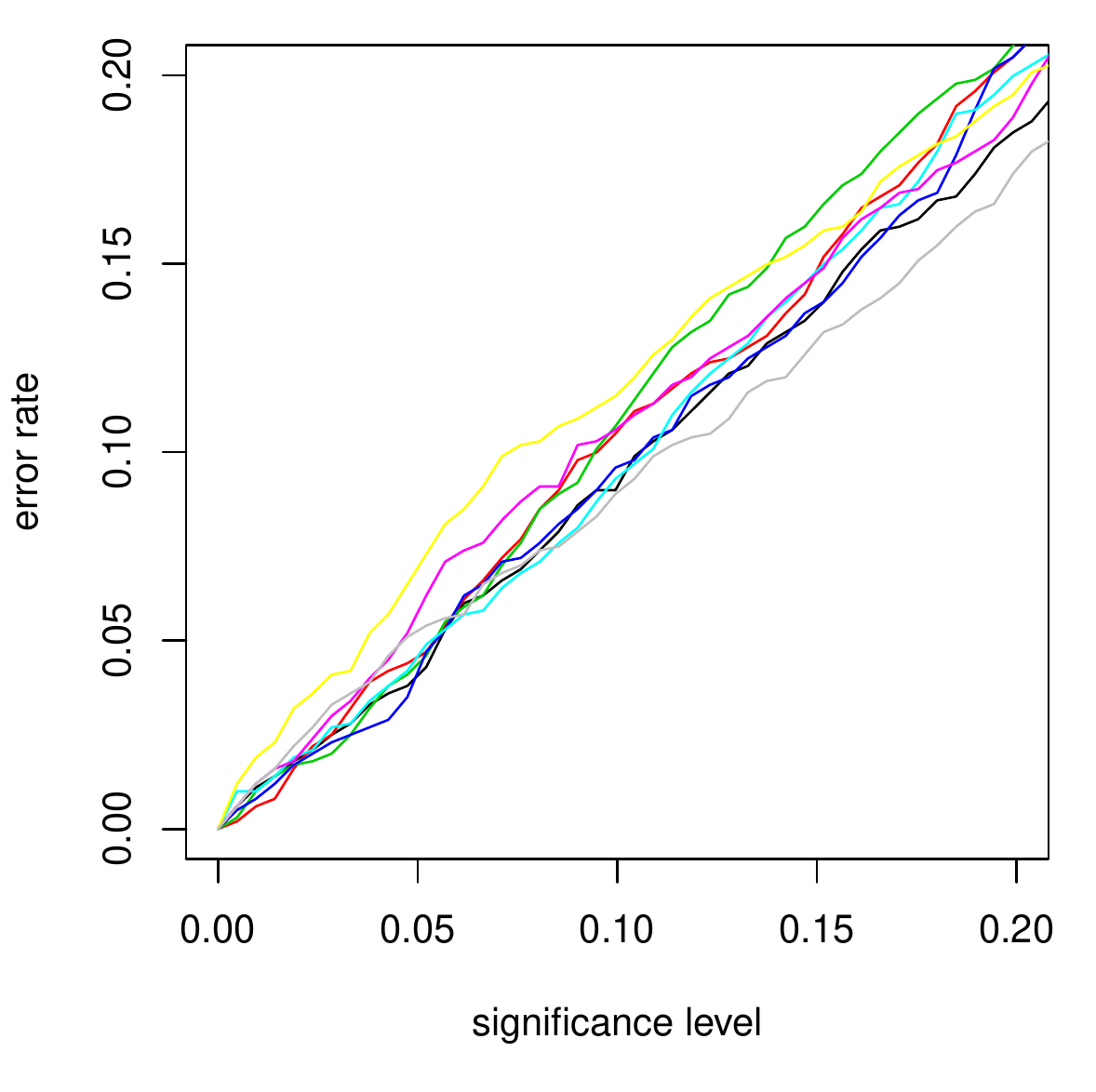}
\end{center}
\caption{Left panels: the calibration plot for the cross-conformal predictor
  with $K=5$ (top) and $K=10$ (bottom) folds
  and the first 8 seeds, 0--7, for the R pseudorandom number generator.
  Right panels: the lower left corner of the corresponding left panel
  (which is the most important part of the calibration plot in applications).}
\label{fig:calibration_CCP2}
\end{figure}

As for the efficiency of the CCP, see Table~\ref{tab:statistics}.
The biggest advantage of the CCP is in the stability of its confidence values:
the standard deviation of the mean confidences is much less than that for the ICP.
However, the CCP also gives higher confidence;
to some degree this can be seen from the table,
but the high variance of the ICP confidence masks it:
e.g., for the first 100 seeds the average of the mean confidence for ICP is $99.16\%$
(with the standard deviation of the mean confidences equal to $0.149\%$,
corresponding to the standard deviation of $0.015\%$ of the average mean confidence).

\begin{table}[tb]
\hspace*{-2.5cm}{\small\begin{tabular}{lrrrrrrrrrr}
  Seed        & 0 & 1 & 2 & 3 & 4 & 5 & 6 & 7 & Average & St.\ dev.\\
  \hline
  mean conf., ICP    & 99.25\% & 99.23\% & 99.00\% & 99.17\% & 99.30\% & 99.12\% & 99.38\% & 99.25\% & 99.21\% & 0.109\%\\
  mean cred., ICP    & 51.31\% & 50.37\% & 49.93\% & 52.45\% & 48.98\% & 50.34\% & 50.18\% & 52.00\% & 50.69\% & 1.074\%\\
  \hline
  mean conf., $K=5$  & 99.22\% & 99.17\% & 99.17\% & 99.24\% & 99.27\% & 99.27\% & 99.30\% & 99.30\% & 99.24\% & 0.050\%\\
  mean cred., $K=5$  & 51.06\% & 49.70\% & 50.26\% & 50.63\% & 49.81\% & 49.42\% & 50.88\% & 51.40\% & 50.39\% & 0.664\%\\
  \hline
  mean conf., $K=10$ & 99.24\% & 99.20\% & 99.20\% & 99.23\% & 99.26\% & 99.28\% & 99.34\% & 99.32\% & 99.26\% & 0.048\%\\
  mean cred., $K=10$ & 51.02\% & 49.69\% & 50.23\% & 50.71\% & 49.70\% & 49.42\% & 50.89\% & 51.39\% & 50.38\% & 0.678\%\\
  \hline
\end{tabular}}
\caption{Mean (over the test set) confidence and credibility for the ICP and the 5-fold and 10-fold CCP.
  The results are given for various values of the seed for the R pseudorandom number generator;
  column ``Average'' gives the average of all the 8 values for the seeds 0--7,
  and column ``St.\ dev.''\ gives the standard deviation of those 8 values.}
\label{tab:statistics}
\end{table}

\section{Conclusion}
\label{sec:conclusion}

Conformal prediction and inductive conformal prediction
are two approaches to the theory of tolerance regions
(see, e.g., \cite{fraser:1957}).
The known validity results for conformal and inductive conformal predictors
can be expressed by saying that they are $1-\epsilon$ expectation tolerance regions,
where $\epsilon$ is the significance level
(see Proposition~\ref{prop:validity-ICP} above for the case of ICPs).
It is also known (\cite{vovk:2012}, Proposition~2a)
that inductive conformal predictors are $1-\delta$ tolerance regions for a proportion $1-\epsilon$
for suitable $\delta$ and $\epsilon$.
On the other hand,
at this time there are no theoretical results about the validity of cross-conformal predictors,
and it is an interesting open problem to establish such results.

\subsection*{Acknowledgments}

The empirical studies described in this paper used the R system 
and the \texttt{gbm} package written by Greg Ridgeway
(based on the work of Freund and Schapire \cite{freund/schapire:1997}
and Friedman \cite{friedman:2001,friedman:2002}).
This work was partially supported by the Cyprus Research Promotion Foundation.

\ifarXiv
  \bibliographystyle{plain}
  \bibliography{%
    /doc/work/r/bib/ait/ait,%
    /doc/work/r/bib/expert/expert,%
    /doc/work/r/bib/games/games,%
    /doc/work/r/bib/general/general,%
    /doc/work/r/bib/math/math,%
    /doc/work/r/bib/prob/prob,%
    /doc/work/r/bib/stat/stat,%
    /doc/work/r/bib/vovk/vovk}
\fi

\ifWP
  \bibliographystyle{plain}
  \bibliography{%
    /doc/work/r/bib/ait/ait,%
    /doc/work/r/bib/expert/expert,%
    /doc/work/r/bib/games/games,%
    /doc/work/r/bib/general/general,%
    /doc/work/r/bib/math/math,%
    /doc/work/r/bib/prob/prob,%
    /doc/work/r/bib/stat/stat,%
    /doc/work/r/bib/vovk/vovk}
\fi

\appendix
\section{An approach based on Fisher's method}
\label{sec:CCP1}

In this appendix we will briefly discuss an approach to cross-conformal prediction
leading to miscalibrated set predictions.

Fisher's method \cite{fisher:1948} of combining p-values $p_1,\ldots,p_K$,
valid when the $K$ p-values are independent,
combines them into one statistic
$
  -2 \sum_{k=1}^K \ln p_k
$
having the chi-squared distribution with $2K$ degrees of freedom.
The corresponding p-value will be denoted $F(p_1,\ldots,p_K)$:
\begin{equation}\label{eq:Fisher}
  F(p_1,\ldots,p_K)
  :=
  \Prob
  \left(
    \chi^2
    \ge
    -2 \sum_{k=1}^K \ln p_k
  \right),
\end{equation}
where $\chi^2$ is a random variable having the chi-squared distribution with $2K$ degrees of freedom.

\emph{Naive cross-conformal predictors} are defined as follows.
The training set is split into $K$ subsets, as in the case of CCPs.
For each $k\in\{1,\ldots,K\}$ find the p-values $p^y_k$ via (\ref{eq:p-CCP-separate}).
Define $p^y:=F(p^y_1,\ldots,p^y_K)$,
and then define confidence, credibility, and set predictors (\ref{eq:ICP}) as before.
In other words,
naive cross-conformal predictors are defined in the same way as cross-conformal predictors
except that the function $F$ is defined by (\ref{eq:Fisher})
rather than by the expression following the $=$ in (\ref{eq:modified-mean}).

Figure~\ref{fig:calibration_CCP1} is the analogue of Figure~\ref{fig:calibration_CCP2}
for naive cross-conformal predictors.
It is obvious that the set predictions are badly miscalibrated;
the p-values computed from different folds are heavily dependent.
This may appear counterintuitive,
but the reader should remember that we are dealing with a somewhat unusual kind of hypothesis testing
in this note (and in the theory of tolerance regions in general):
instead of testing some properties of the data-generating distribution
we are testing hypotheses about data.

We do not give the efficiency results
(such as those given in Table~\ref{tab:statistics})
for the naive CCP
since efficiency without validity is meaningless.

\begin{figure}[tb]
\begin{center}
  \includegraphics[width=0.45\textwidth]{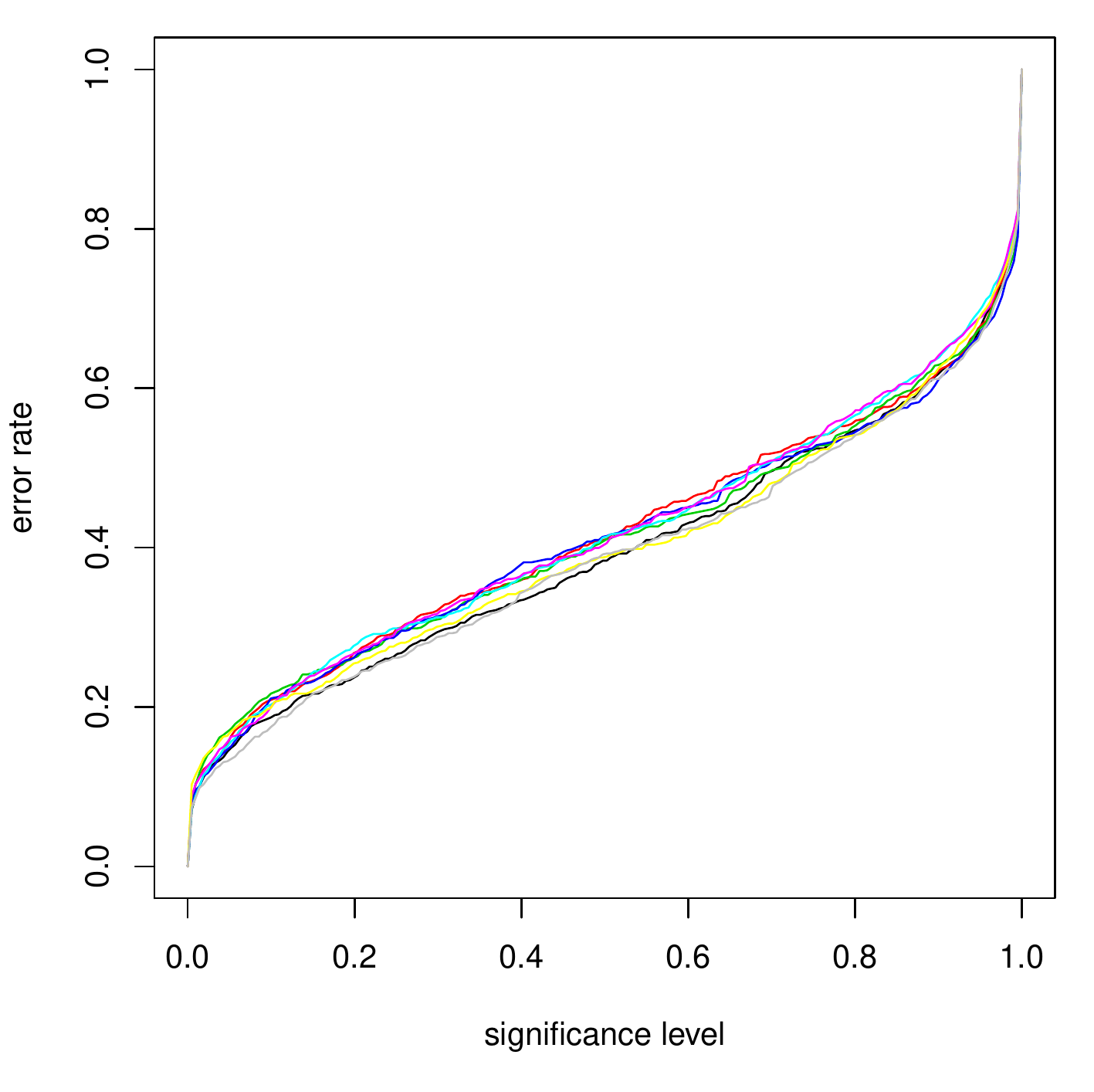}
  \includegraphics[width=0.45\textwidth]{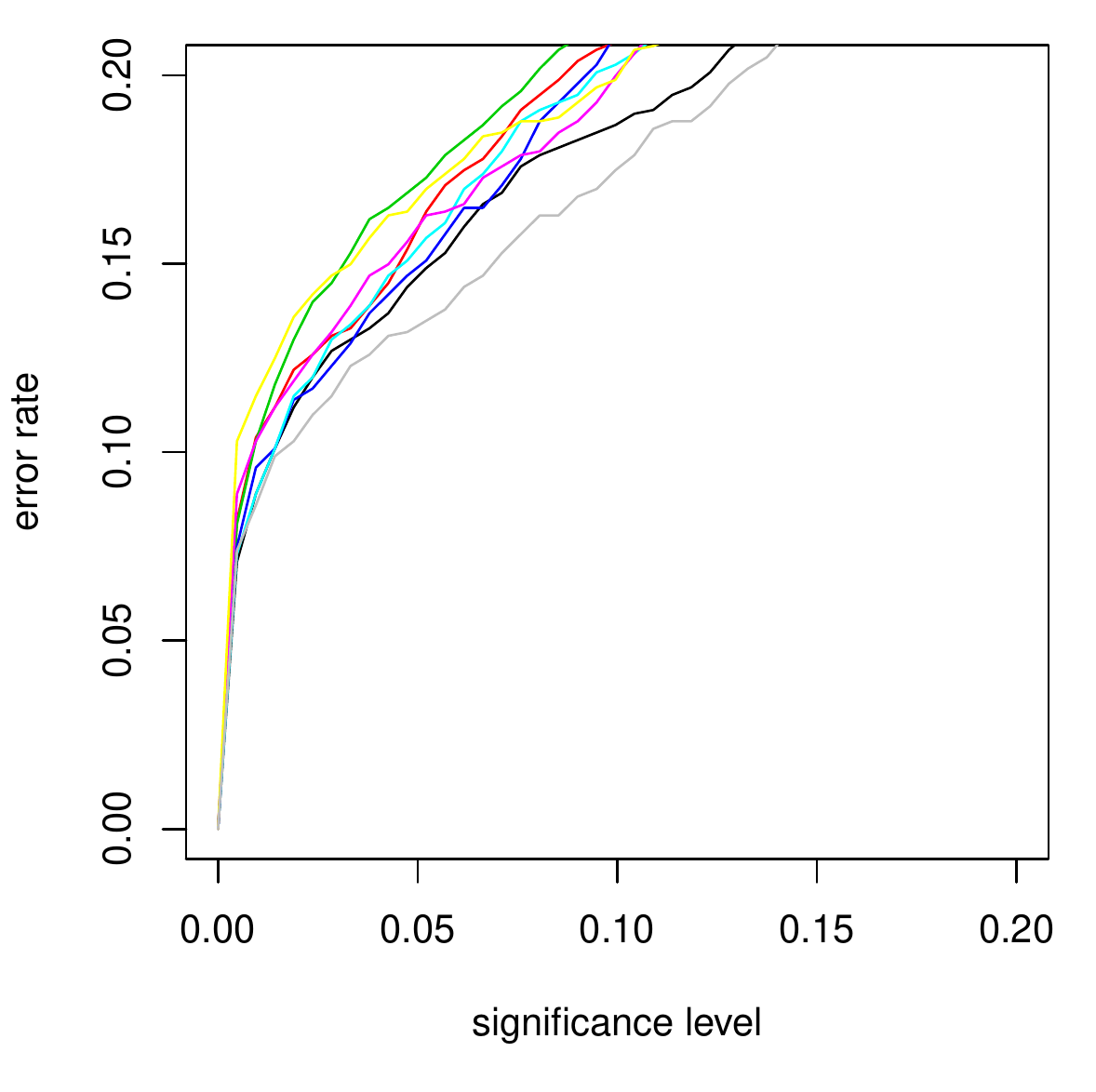}\\[5mm]
  \includegraphics[width=0.45\textwidth]{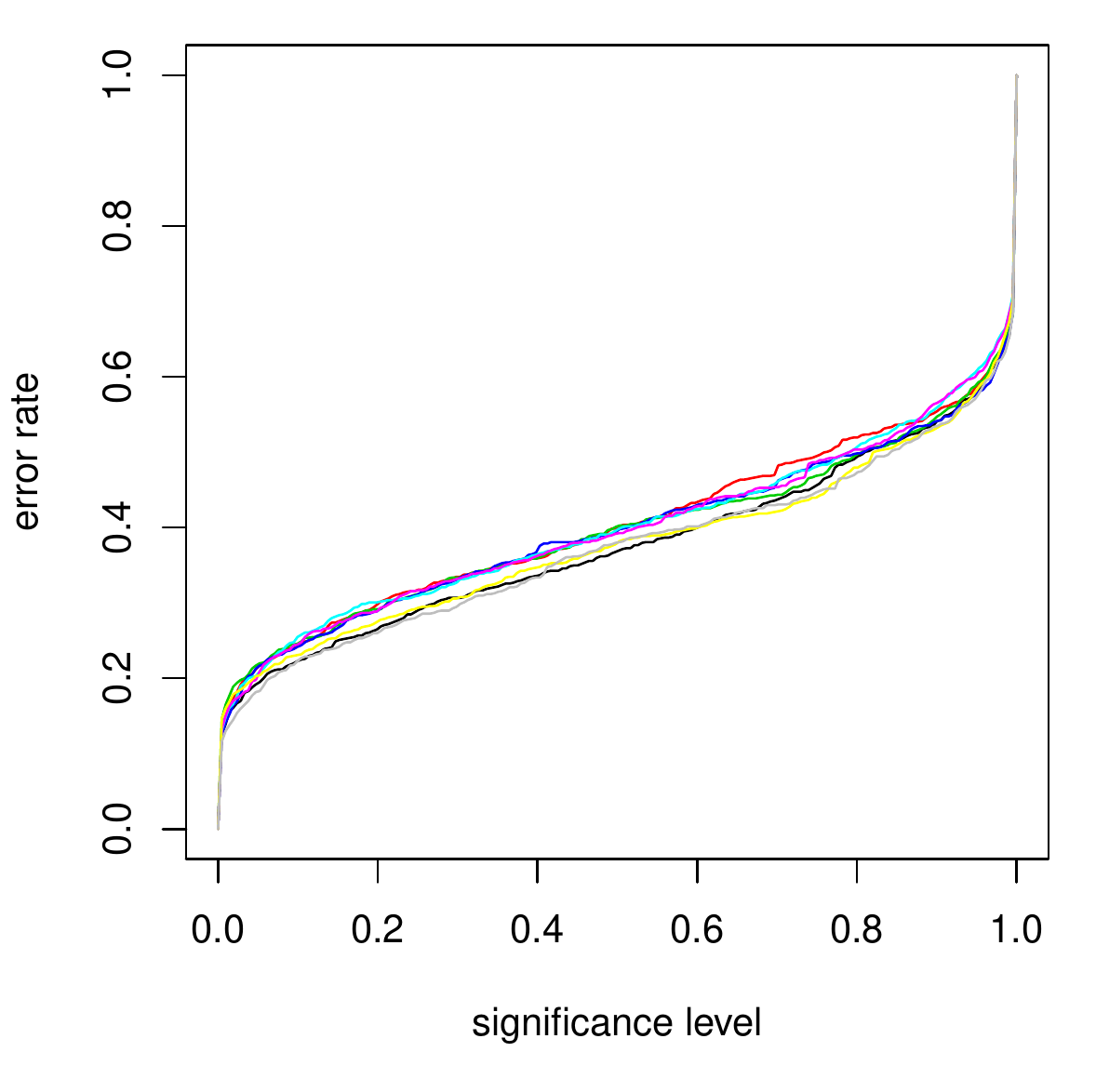}
  \includegraphics[width=0.45\textwidth]{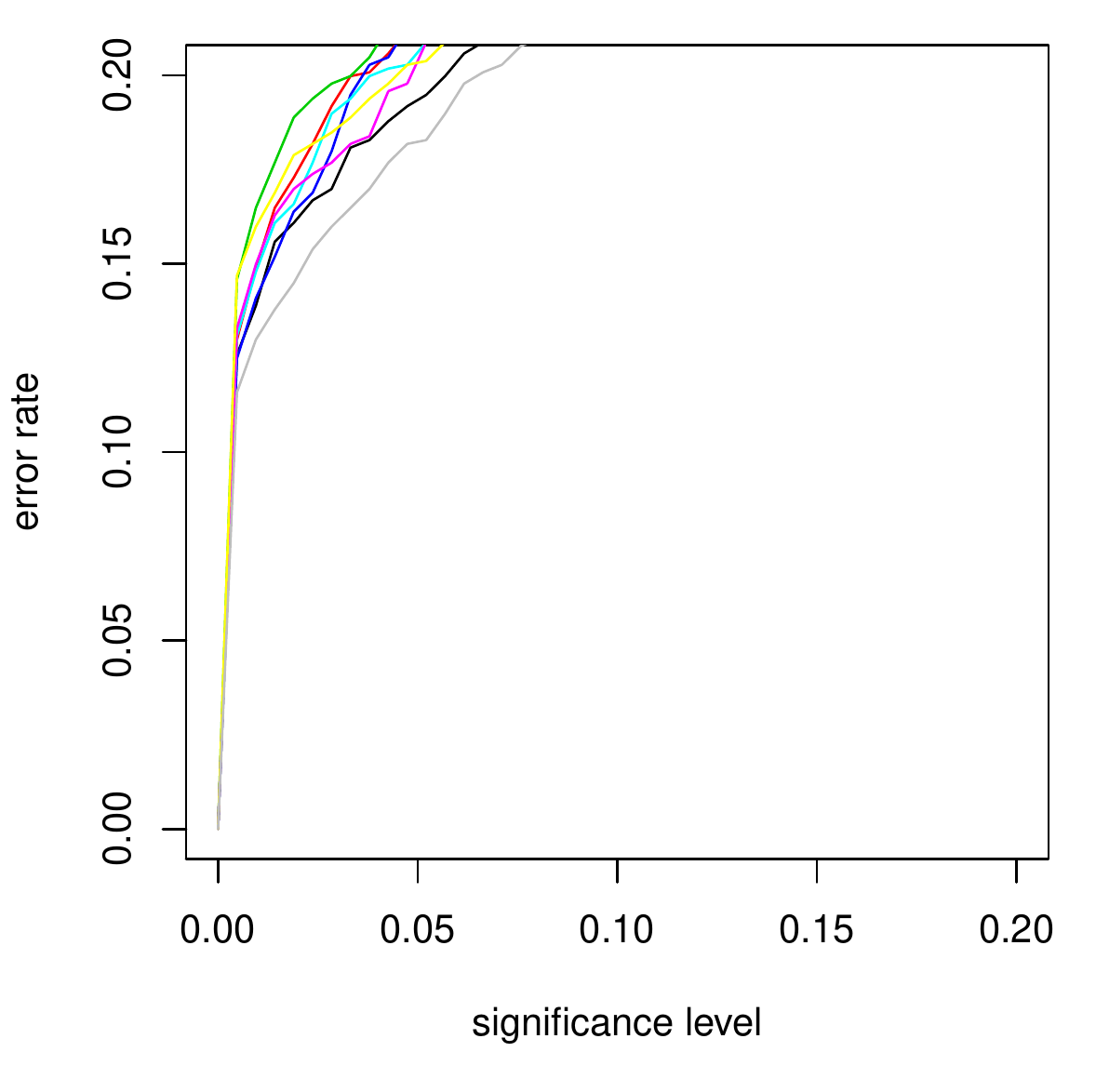}
\end{center}
\caption{The analogue of Figure~\ref{fig:calibration_CCP2} for the naive cross-conformal predictor.}
\label{fig:calibration_CCP1}
\end{figure}
\end{document}